\documentclass[twocolumn, switch, tikz]{article} 

\usepackage{preprint}

\usepackage{amsmath, amsthm, amssymb, amsfonts}

\usepackage[numbers,square]{natbib}
\usepackage[utf8]{inputenc}	
\usepackage[T1]{fontenc}	
\usepackage{xcolor}		
\usepackage[colorlinks = true,
            linkcolor = purple,
            urlcolor  = blue,
            citecolor = cyan,
            anchorcolor = black]{hyperref}	

\usepackage{lineno}		
\usepackage{float}			
\usepackage{todonotes}
\usepackage{preprint}
\usepackage[utf8]{inputenc} 
\usepackage[T1]{fontenc}    
\usepackage{hyperref}       
\usepackage{url}            
\usepackage{booktabs}       
\usepackage{amsfonts}       
\usepackage{nicefrac}       
\usepackage{microtype}      
\usepackage{lipsum}
\usepackage{fancyhdr}       
\usepackage{graphicx}       
\graphicspath{{media/}}     
\usepackage{subfig}
\usepackage{threeparttable}
\usepackage{array}
\usepackage{makecell}
\usepackage{multirow}
\usepackage{graphicx}
\usepackage{booktabs}
\usepackage{todonotes}

 \usepackage{lipsum}		

\usepackage{newfloat}
\DeclareFloatingEnvironment[name={Supplementary Figure}]{suppfigure}
\usepackage{sidecap}
\sidecaptionvpos{figure}{c}

\usepackage{titlesec}
\titlespacing\section{0pt}{12pt plus 3pt minus 3pt}{1pt plus 1pt minus 1pt}
\titlespacing\subsection{0pt}{10pt plus 3pt minus 3pt}{1pt plus 1pt minus 1pt}
\titlespacing\subsubsection{0pt}{8pt plus 3pt minus 3pt}{1pt plus 1pt minus 1pt}

\usepackage{tikz,xcolor,hyperref}

\usepackage{import}
\usetikzlibrary{positioning}
\usetikzlibrary{3d} 
\usepackage{multicol}

\title{Network architecture search of X-ray based scientific applications}

\usepackage{authblk}

\author[1]{Adarsha Balaji}
\author[2]{Ramyad Hadidi}
\author[2]{Gregory Kollmer}
\author[2]{Mohammed E. Fouda}
\author[1,3]{Prasanna Balaprakash}

\affil[1]{Argonne National Laboratory, Chicago, IL, USA.}
\affil[2]{Rain Neuromorphics Inc., San Francisco, CA, USA.}
\affil[3]{Oak Ridge National Laboratory, Oak Ridge, TN, USA}

\begin{document}

\twocolumn[ 
  \begin{@twocolumnfalse} 
  
\maketitle

\begin{abstract}
X-ray and electron diffraction-based microscopy use bragg peak detection and ptychography to perform 3-D imaging at an atomic resolution. Typically, these techniques are implemented using computationally complex tasks such as a Psuedo-Voigt function or solving a complex inverse problem. Recently, the use of deep neural networks has improved the existing state-of-the-art approaches. However, the design and development of the neural network models depends on time and labor intensive tuning of the model by application experts. To that end, we propose a hyperparameter (HPS) and neural architecture search (NAS) approach to automate the design and optimization of the neural network models for model size, energy consumption and throughput. We demonstrate the improved performance of the auto-tuned models when compared to the manually tuned BraggNN and PtychoNN benchmark. We study and demonstrate the importance of the exploring the search space of tunable hyperparameters in enhancing the performance of bragg peak detection and ptychographic reconstruction. Our NAS and HPS of (1) BraggNN achieves a 31.03\% improvement in bragg peak detection accuracy with a 87.57\% reduction in model size, and (2) PtychoNN  achieves a 16.77\% improvement in model accuracy and a 12.82\% reduction in model size when compared to the baseline PtychoNN model. When inferred on the Orin-AGX platform, the optimized Braggnn and Ptychonn models demonstrate a 10.51\% and 9.47\% reduction in inference latency and a 44.18\% and 15.34\% reduction in energy consumption when compared to their respective baselines, when inferred in the Orin-AGX edge platform.




\end{abstract}
\vspace{0.35cm}

  \end{@twocolumnfalse} 
] 



\section{Introduction}\label{sec:intro}

X-ray and electron diffraction microscopy has emerged as a key nano-scale imaging technique used in the design and study of advanced materials for a variety of scientific fields, such as semiconductor device fabrication \cite{deng2017nanoscale}, cell biology \cite{hemonnot2017imaging} and battery technology \cite{tsai2019correlated}. 
High energy characterization techniques, such as high energy diffraction microscopy (HEDM) \cite{park2017far}, bragg coherent diffraction imaging \cite{huang2012three} and Lorentz transmission electron microscopy (LTEM), have been developed over the past decade to achieve this. In these techniques, the imaged nano-scale object is illuminated with a coherent beam-line from an X-ray source and the scattered intensities are measured in the far field. The \textit{amplitude} and \textit{phase} information, required to generate the 3-D image of the object, is recovered using ptychographic techniques to invert the coherent diffraction images. The inversion of the measured image and the bragg peak detection require solving a complex inverse problem and the Pseudo-Voigt function, respectively, which are computationally expensive and traditionally performed offline.


Deep Neural Networks (DNNs) have shown remarkable success in accelerating the detection of bragg peaks, as demonstrated by BraggNN \cite{liu2022braggnn}, and in facilitating the ptychographic reconstruction of 3-D images of observed objects, as achieved by PtychoNN \cite{cherukara2020ai}. Typically, the experimental process involves acquiring data from the detector and then transferring this data to central storage located away from the detector. This is followed by the development of predictive models for bragg peak analysis and ptychographic reconstruction. During deployment, the need for real-time processing and handling high-volume data acquisitions poses significant challenges in allocating adequate resources for the transmission and remote processing of raw data. Implementing edge processing—analyzing the data close to its point of capture—enables the creation of optimized processing pipelines that efficiently overcome these limitations. Such models are ideally suited for deployment on GPU-based edge devices, like those in the high-end Nvidia Jetson series. Nonetheless, ensuring the deployability of DNN models on energy and resource-constrained hardware necessitates a thorough investigation into the neural network architecture and the optimization of performance-related hyperparameters.


Neural architecture search (NAS) methods are designed to automatically search for the best performing network architecture for a given dataset, within predetermined constrains. In addition, hyperparameter search (HPS) methods help find the performance optimizing hyperparameter settings for a chosen network architecture. In particular, these methods use search algorithms, like evolutionary algorithms and bayesian optimization, to explore a user-constrained search space and choose a network architecture and its network hyperparameters. Hyperparameter optimization frameworks, such as Autotune \cite{koch2018autotune}, Optuna \cite{akiba2019optuna} and Deephyper \cite{balaprakash2018deephyper}, have shown to out-perform manually designed and tuned neural architectures for image classification, natural language processing and other scientific applications.  

In this work, we develop NAS and HPS for DNNs used in bragg peak detection and ptychographic reconstruction. We define the NAS and HPS search space for the DNN to explore feed forward and deep convolutional architectures operators. We perform the NAS and HPS search using Deephyper \cite{balaprakash2018deephyper} - an open-source, scalable automated machine learning (AutoML) package and contrast its performance with state-of-the-art packages like Optuna \cite{akiba2019optuna}.










The contributions of this work are summarized as follows: 

\begin{itemize}
\item We demonstrate the architecture optimization of two scientific applications with Deephyper (HPS+NAS). Our results show the Pareto optimal graphs illustrating the relation between model size and accuracy. We show a decrease the number of trainable parameters by 87.57\% and 12.67\% for BraggNN and PtychoNN model while achieving better accuracy.

\item We highlight the improved performance of the Deehyper framework against state-of-the-art HPS and NAS tools (Optuna) in terms of convergence speed. 

\item We perform hardware evaluation for inferring the optimized models using Orin-AGX, an Nvidia GPU based edge device, and compare the latency and energy consumption of the baseline vs the optimized models. When inferred on the Orin-AGX platform, demonstrate a 10.51\% and 9.47\% reduction in latency and a 44.18\% and 15.34\% reduction in energy consumption when compared to their respective baselines, when inferred in the Orin-AGX edge platform.


    
\end{itemize}

\begin{figure}[t!]
	\centering
\includegraphics[width=1\columnwidth]{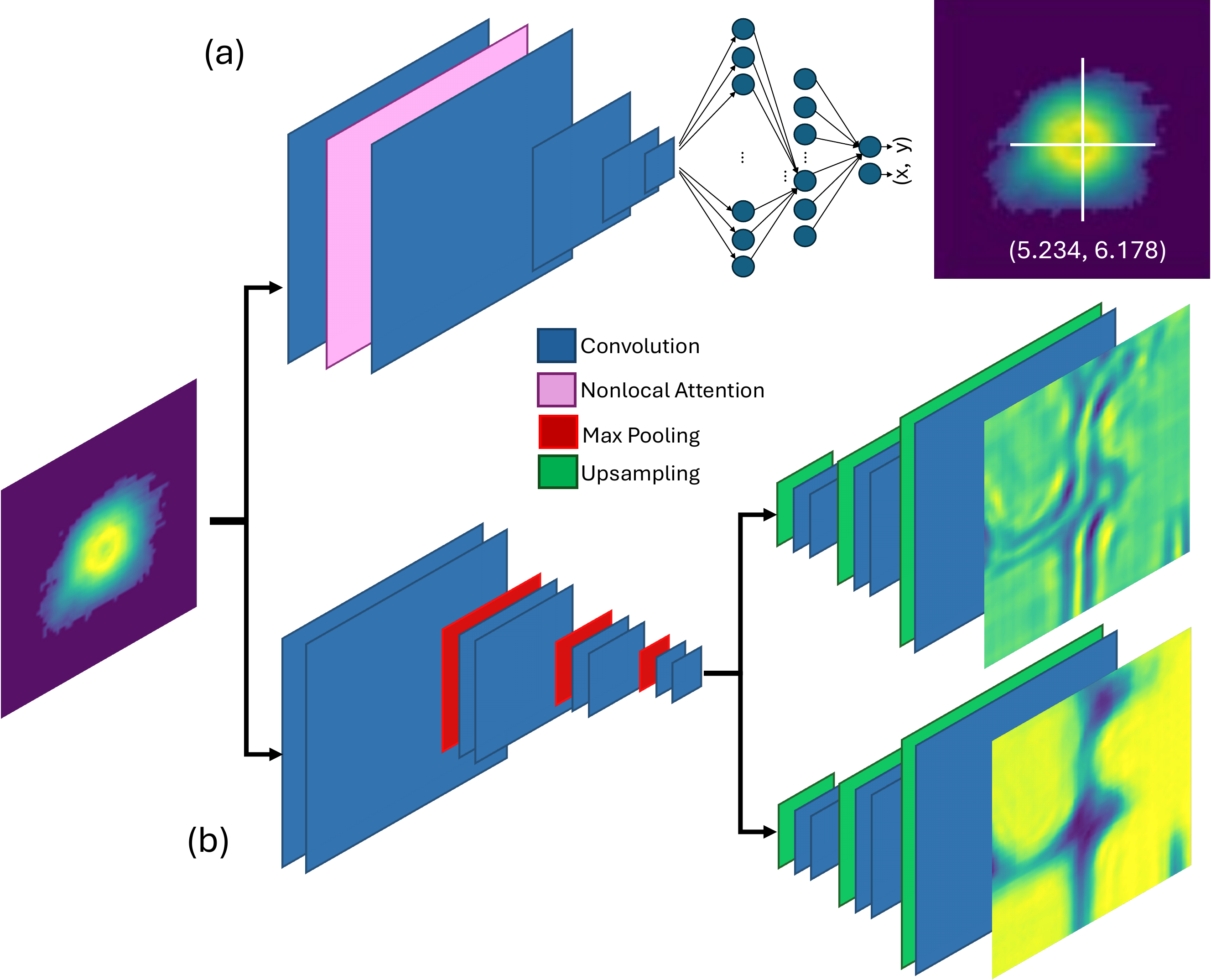}
	\caption{Neural Network models used for X-ray scientific applications; (a)BraggNN and (b) PtychoNN models.}
	\label{fig:baseline-functional} 
\end{figure}

\section{Background}\label{sec:background}

\subsection{BraggNN}\label{subsec:braggnn}
X-ray diffraction-based microscopy using HEDM techniques provide information on the 3-D structure of polycrystalline materials. A typical HEDM experiment involves illuminating a sample with X-rays and capturing between 1440-3600 diffraction images, while rotating the material at a constant speed. The reconstruction of the Far-Field (FF) HEDM data involves the detection of Bragg peak position at a sub-pixel accuracy. 

The BraggNN model \cite{liu2022braggnn} is a deep learning model designed to analyze the fast bragg peaks generated in an X-ray diffraction based microscopy experiment. The model is trained to precisely localize the position of a bragg peak for a HEDM experiment \cite{park2017far} designed to provide information on the 3-D structure and evolution of poly-crystalline materials in scientific applications. The task of detecting the bragg peak is expressed as a regression problem to predict the X and Y coordinates of the bragg peak. Figure \ref{fig:baseline-functional} (a) demonstrates the baseline model of BraggNN. The model architecture is designed using a series of convolution layers to perform feature extraction, followed by a series of dense fully connected layers to predict the co-ordinates of the bragg peaks. The input to the model is a 11$\times$11$\times$1 patch of bragg peak pixels generated from the experiment and the performance of the model is measured as the accuracy of its estimated bragg peak position (X,Y) when compared the traditional pseudo-Voigt fit. The authors claim an accuracy of 0.29 and 0.57 pixels for 75\% and 95\% of the peaks, respectively, when compared to the pseudo-Voigt method. 

\subsection{PtychoNN} \label{subsec:ptychonn}
Ptychographic imaging is performed by scanning a coherent X-ray or electron beam across the sample while measuring the scattered intensities in the far field to create a diffraction image. The sample image is recovered by algorithmically inverting the measured coherent diffraction images which in-turn requires the solution to a complex inverse problem to generate the phase and amplitude information for every pixel in the image. 

Figure \ref{fig:baseline-functional} (b) illustrates the baseline model. The PtychoNN model \cite{cherukara2020ai} is designed to take an input of raw X-ray scaterring data and output a reconstruction of the sample using amplitude and phase information. The model architecture proposed by the authors consists of an encoder arm to learn a representation of the input and two decoder arms, to predict the real-space amplitude and phase information from the input. The encoder arm is built using convolutional and max-pooling layers to extract the latent representations from the given input, and the decoder arms are built using convolutional and up-sampling layers to reconstruct the real-space amplitude and the phase information from the extracted latent representations. The authors claim that the PtychoNN model performs the amplitude and phase retrieval 300 $\times$ faster than the state-of-the-art iterative phase retrieval solution.


\begin{figure}[t!]
\centering
\includegraphics[width=0.75\columnwidth]{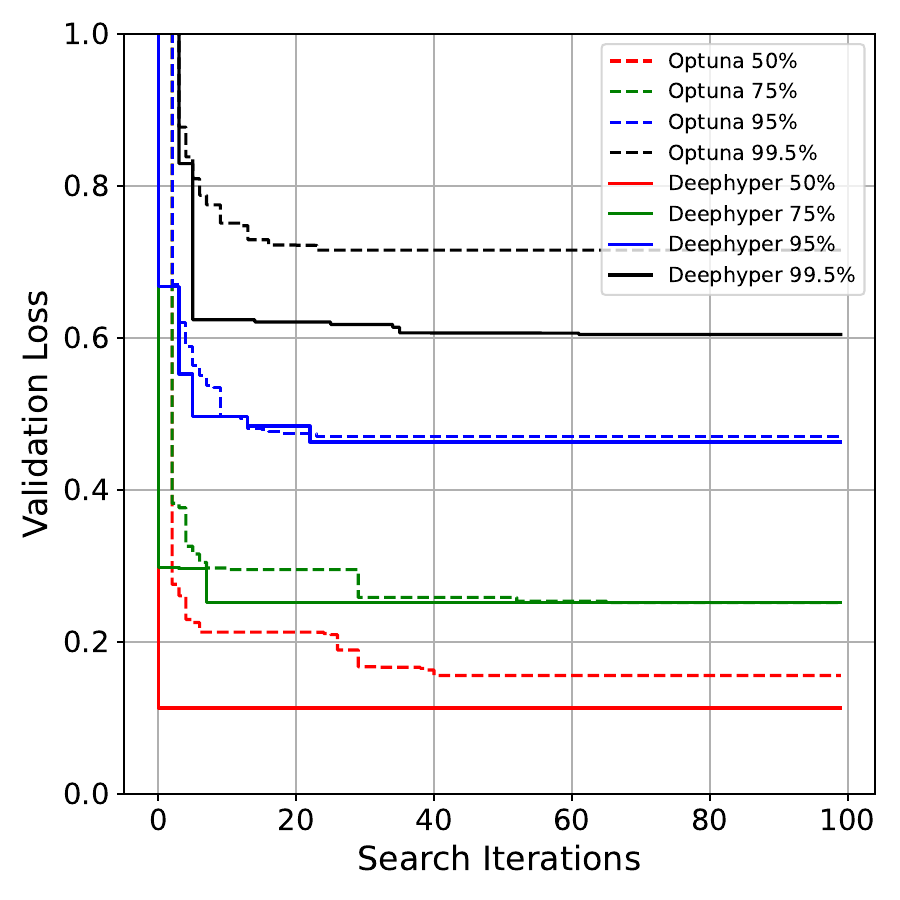}
\caption{BraggNN: Deephyper vs Optuna - Convergence speed.}
	\vspace{-5pt}
	\label{fig:mo-deepvsoptuna}
\end{figure}

\subsection{Search Methods}
The widespread success of deep neural networks (DNNs) has created a need for architecture exploration, where users manually design increasingly complex neural architectures. This has led to an increased interest in AutoML \cite{yao2018taking} as a method to explore neural architecture search (NAS) and hyperparameter optimization (HPS), with an aim to find the best network architecture and its respective hyperparameters for any given learning task and dataset. The most naive HPS approaches are a brute-force and random search methods, in which the user either specifies a finite set of values for each hyperparameter and randomly samples a hyperparameter configuration within the constraints of a budget. As expected these methods are not scalable and inefficient at exploring the search space.

To address this, hyperparameter optimization methods that perform a guided search of the neural search space by iteratively generating new hyperparameter configurations based on the performance of a prior configurations have been explored. Bio-inspired algorithms like Particle swarm optimization \cite{escalante2009particle} and evolutionary algorithms \cite{back1993overview} are efficient at initializing a population of hyperparameter configurations and iteratively moving the population towards the best local and global solutions in the search space. However, in recent years, Bayesian optimization has emerged as the state-of-the-art optimization framework for AutoML systems. Bayesian optimization is a probabilistic, iterative algorithm with two main components: a surrogate model and an acquisition function. Bayesian optimization builds a probabilistic surrogate model, usually in the form of a Gaussian process or a tree-based model, which is used to map the different hyperparameter configurations to their performance with some measure of uncertainty. 

Several hyperparameter search toolkits exits to tune deep neural network architectures. These include MENNDL \cite{young2015optimizing}, an evolutionary optimization approach;  Hyperband \cite{li2018hyperband}, which aims at accelerating random search by implementing early stopping criteria; Optuna \cite{akiba2019optuna} and Hyperopt \cite{bergstra2015hyperopt}, which both offer random sampling and Tree-structured Parzen Estimator (TPE) methods; and DeepHyper \cite{balaprakash2018deephyper}, which is based on Bayesian optimization methods.

\begin{table}[t!]
	\renewcommand{\arraystretch}{1}
	\setlength{\tabcolsep}{2pt}
	\caption{Search space considered to optimize the baseline.}
    \vspace{10pt}
	\centering
    \label{tab:exp1_var}
	\begin{threeparttable}
	{\fontsize{8}{10}\selectfont
		\begin{tabular}{c|l}
			\hline
			\textbf{Application} & \textbf{Hyperparameters}\\
			\hline 
            mlpBragg & epochs (nepochs) $\in$ [1, 1000] \\
             & number of units per layer (nunits) $\in$ [1,1000] \\
             & number of hidden layers (nhidden) $\in$ [1,10] \\
             & batch size (batch) $\in$ [8, 16, 32, 64] \\
             & learning rate (lr) $\in$ [0.001, 1] \\
    		\hline 

            cnnBragg & epochs (nepochs) $\in$ [1, 1000] \\
             & batch size (batch) $\in$ [8, 16, 32, 64] \\
             & learning rate (lr) $\in$ [0.001, 1] \\
             & number of filters (nfilters) $\in$ [1,256] \\
             & number of convolution layers (n`conv) $\in$ [1,128] \\
            \hline
            mlpPtycho & epochs (nepochs) $\in$ [1, 1000] \\
             & number of units per layer (nunits) $\in$ [1,1000] \\
             & number of hidden layers (nhidden) $\in$ [1,10] \\
             & batch size (batch) $\in$ [8, 16, 32, 64] \\
             & learning rate (lr) $\in$ [0.001, 1] \\     

            \hline
            cnnPtycho & epochs (nepochs) $\in$ [1, 1000] \\
             & batch size (batch) $\in$ [8, 16, 32, 64] \\
             & learning rate (lr) $\in$ [0.001, 1] \\
             & number of filters (nfilters) $\in$ [1,256] \\
             & number of encoder convolution layers (enconv) $\in$ [1,128] \\
             & number of decoder-1 upsample layers (deconv-1) $\in$ [1,128] \\
             & number of decoder-2 upsample layers (deconv-2) $\in$ [1,128] 
            \vspace{0pt}
        \end{tabular}}
	\end{threeparttable}
\end{table}

For this work, we use Deephyper \cite{balaprakash2018deephyper} to perform HPS and NAS. Figure \ref{fig:mo-deepvsoptuna} compares the convergence speed between Deephyper and Optuna, a state-of-the-art HPS tool, for a multi-objective optimization problem. In the experiment, the baseline BraggNN model is optimized for four objectives. We run the HPS for 100 iterations. With the identical initial conditions Deephyper outperforms the state-of-the-art Optuna on two key metrics: (1) convergence speed - Deephyper consistently finds an optimal set of hyper-parameters with fewer search iterations when compared to Optuna, and (2) performance - the optimized models generated using Deephyper outperform the Optuna models on all objectives.

\section{Methodology} 
\subsection{Neural Architecture Search}
The NAS approach we propose involves: (1) a search space that defines a multi-layer perceptron architecture, (2) a search space that defines a deep convolution architecture, (3) a search method that will explore the architecture search space, and (4) an evaluation method that will measure the performance of the network, in terms of accuracy, confidence and model size.

\subsubsection{BraggNN Architecture Search Space}
The detailed implementation of the baseline BraggNN architecture is described in Section \ref{subsec:braggnn}. We extensively search for an MLP-based implementation, mlpBragg, and CNN-based implementation, cnnBragg, of the feature extraction block of BraggNN. The architecture hyperparameters for mlpBragg and cnnBragg are presented in Table \ref{tab:exp1_var}. 

The mlpBragg model is composed of a series of \text{m} fully connected layers of ReLU neurons 
and the number of neurons per layer (nunits) and the number of sequential layers (nhidden) are parameterized. The feature extraction block of cnnBragg is composed of a series of convolution layers acting as the feature extractor. The number of convolution layers (enconv), upsample layers (deconv-1, deconv-2) and the number of filters (nfilters) in each layer are parameterized.

\begin{figure}[t!]
	\centering
	\centerline{\includegraphics[height=0.99\columnwidth, width=0.99\columnwidth]{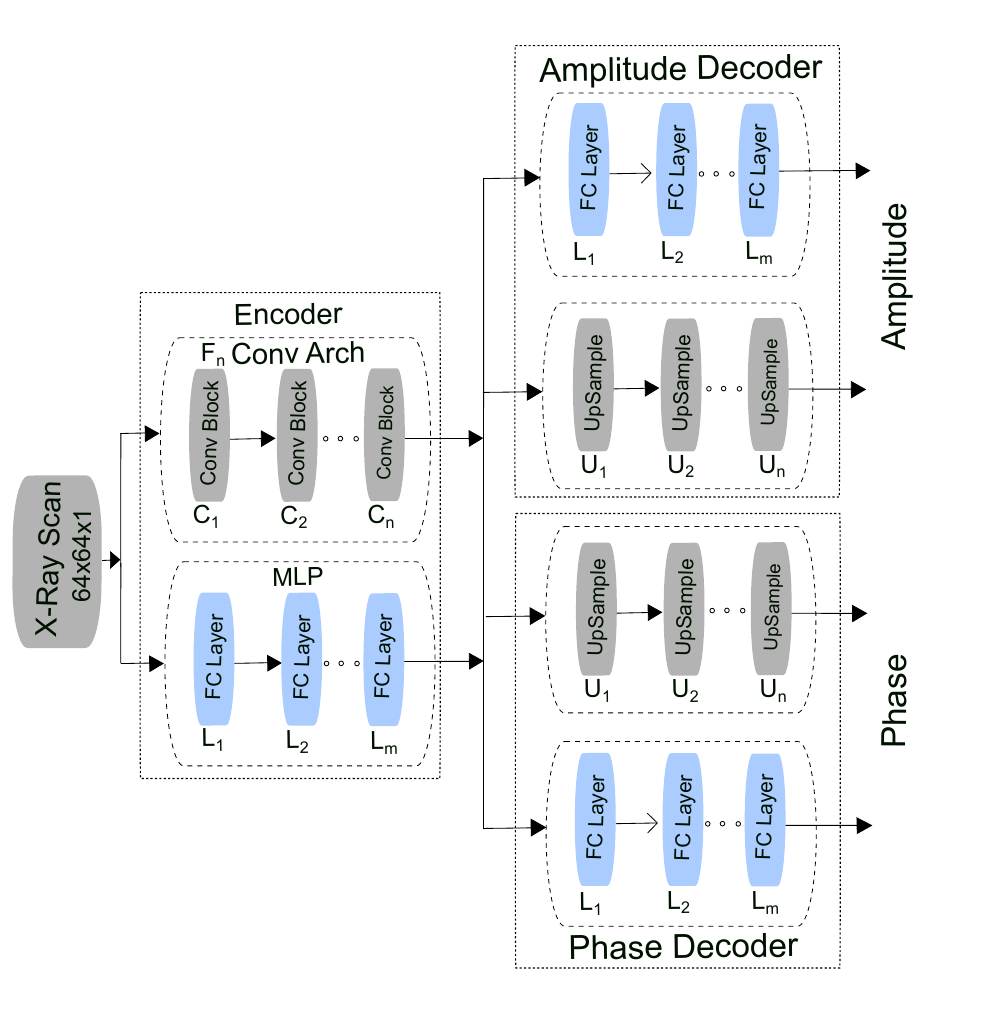}}
	\caption{Proposed neural architecture search of the PtychoNN model.}
	\label{fig:ptychonn-nas}
\end{figure}

\subsubsection{PtychoNN Architecture Search Space}
Figure \ref{fig:ptychonn-nas} illustrates a baseline PtychoNN architecture, indicated in gray. The detailed implementation of the PtychoNN architecture is described in Section \ref{subsec:ptychonn}. We extensively search (1) a MLP-based implementation - mlpPtycho, indicated in blue, and (2) a CNN-based implementation - cnnPtycho, based on the encoder and decoder architecture of PtychoNN, indicated in grey in Figure \ref{fig:ptychonn-nas}. The architecture hyperparameters for mlpPtycho and cnnPtycho are presented in Table \ref{tab:exp1_var}.

The mlpPtycho is composed of a series of \textit{m} fully connected (FC) layers with \textit{nunits} ReLU neurons per layer. The number of layers (m) and the number of neurons per layer (nunits) are parameterized. For cnnPtycho, the encoder block is composed of a series of convolution layers acting as the feature extractor and the decoder blocks are composed of a series of up-sample layers. The number of convolution layers (n) and the number of filters (nfilters) in each layer are parameterized.

\subsection{Hyper-parameter Tuning}       
To extensively explore the search space of all or a subset of the hyperparameters during training, we instantiate a hyperparameter search module within DeepHyper \cite{balaprakash2018deephyper}, an open source software package.

DeepHyper performs a hyperparameter search by repeating the following steps: (1) \textit{sampling} a promising hyperparameter configuration, (2) \textit{evaluating} the sampled hyperparameter configuration, and finally (3) \textit{updating} the objective or a set of objectives, used to bias a future sampling of the hyperparameter space. For a hyperparameter search for a multi-objective optimization problem, Deephyper combines the set of objectives into a scalar using a Chebyshev scalarization function \cite{giagkiozis2015methods}, The DeepHyper evaluator performs a hyperparameter search using a custom asynchronous model-based search (AMBS) using Baysian Optimization.

In this work, the DeepHyper problem can be configured as $\chi = (\chi_A, \chi_P)$, where $\chi_A$ represents the topology of the neural network model and $\chi_P = (\chi_c, \chi_d, \chi_n)$ defines the continuous ($\chi_c$) (eg. learning rate, training epochs), discrete ($\chi_d$) and non-ordinal ($\chi_n$) hyperparameters.

The problem of finding the hyperparameters can be formulated as a single-objective optimization problem: 

\vspace{-2pt}
\begin{equation}
    \label{eq:opt1}
    \text{minimize} \:  err_\nu([\chi_A,\chi_P])
\end{equation}
where $err_\nu$ is the validation dataset error. The problem can also be formulated as a multi-objective optimization problem: 

\vspace{-2pt}
\begin{equation}
    \label{eq:opt2}
    \text{minimize} \:  (err_{\nu}([\chi_A, \chi_P]), \epsilon_\nu([\chi_A, \chi_P]) ) 
\end{equation}
where $\epsilon_{\nu}([\chi_A, \chi_P])$ are the neural network architecture related objective, such as number of hidden layers and number of neurons per hidden layer.

The DeepHyper package also supports synchronous search methods such as evolutionary computation (EC) \cite{fortin2012deap} and hyperband \cite{Li2016HyperbandAN} and asynchronous search techniques such as random search (RS).

\section{Model Optimization} \label{sec:modelopt}

\subsection{Search Space}
Table \ref{tab:exp1_var} shows the search space for the hyperparameters explored using the Deephyper framework \cite{balaprakash2018deephyper}. These include the learning rate ($\alpha$), the batch size ($batch$) and the number of epochs ($nepochs$). The performance of optimized models is quantitatively evaluated for the following metrics: 

\subsubsection{BraggNN}
\textit{Accuracy}: The model accuracy is measured as the mean-square error (MSE) between the position (X,Y) of the Bragg peak vs peak position obtained from the conventional pseudo-Voigt fit.

\textit{Quantile loss}: the quantile statistic is used as a confidence metric to summarize the relative location of Bragg peak prediction (X,Y) independently from a particular underlying probability distribution of the inference dataset. In this case, the model is fitted by minimizing the quantile loss, which can be defined as the asymmetrically weighted sum of absolute deviations.

\subsubsection{PtychoNN}
\textit{Accuracy}: The model accuracy is measured as the mean absolute error (MAE) of the PtychoNN generated real-space amplitude and phase pixel-pixel values when compared to Ptycholib.  

\textit{Model Size}: The model size is measured as the number of trainable parameters in the model. A reduction in model size could significantly reduce the inference time of the models.


\section{Experimental Setup}\label{sec:exp}

In this section, we evaluate the HPS and NAS of the BraggNN and PtychoNN models. We use the DeepHyper framework \cite{balaprakash2018deephyper} to perform the hyperparameter search for an application model optimized for (1) a single-objective, such as application accuracy, as seen in their respective baseline models, and (2) multiple-objective, such as application accuracy, model confidence and model size, to ensure optimized inference on a edge platform. Each evaluation consists of training a generated network and its hyperparameters and optimizing the generated network model for the objectives described in Section \ref{sec:modelopt}.

\begin{figure*}[t!]
	\centering
	\centerline{\includegraphics[height=0.85\columnwidth, width=0.95\textwidth]{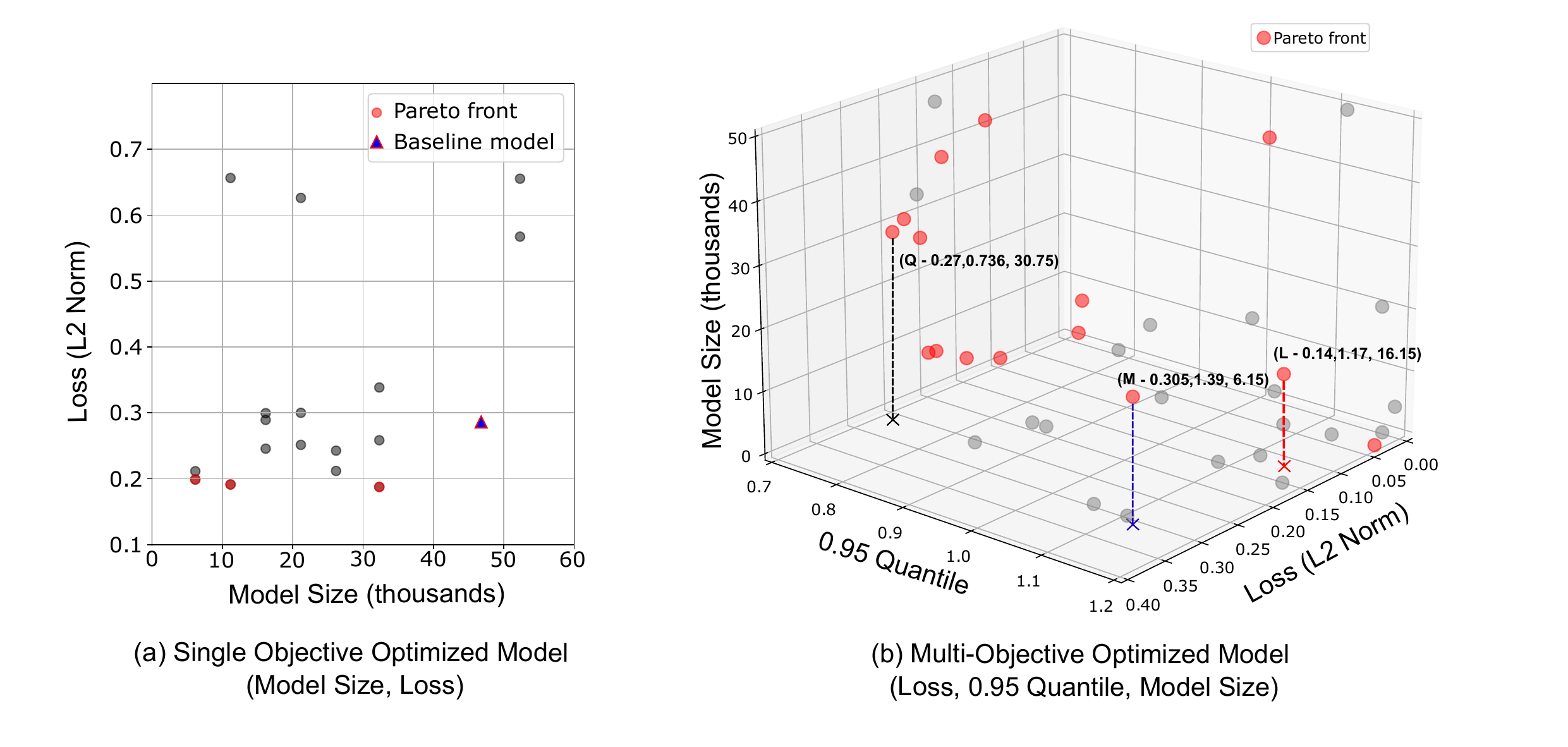}}
	\caption{BraggNN Model Optimization: pareto front of (a) single objective - model accuracy and (b) multi-objective optimization, respectively}
	\label{fig:braggsw-pareto}
\end{figure*}

\begin{figure*}[t!]
	\centering
	\vspace{-25pt}
	\centerline{\includegraphics[height=0.8\columnwidth, width=0.7\textwidth]{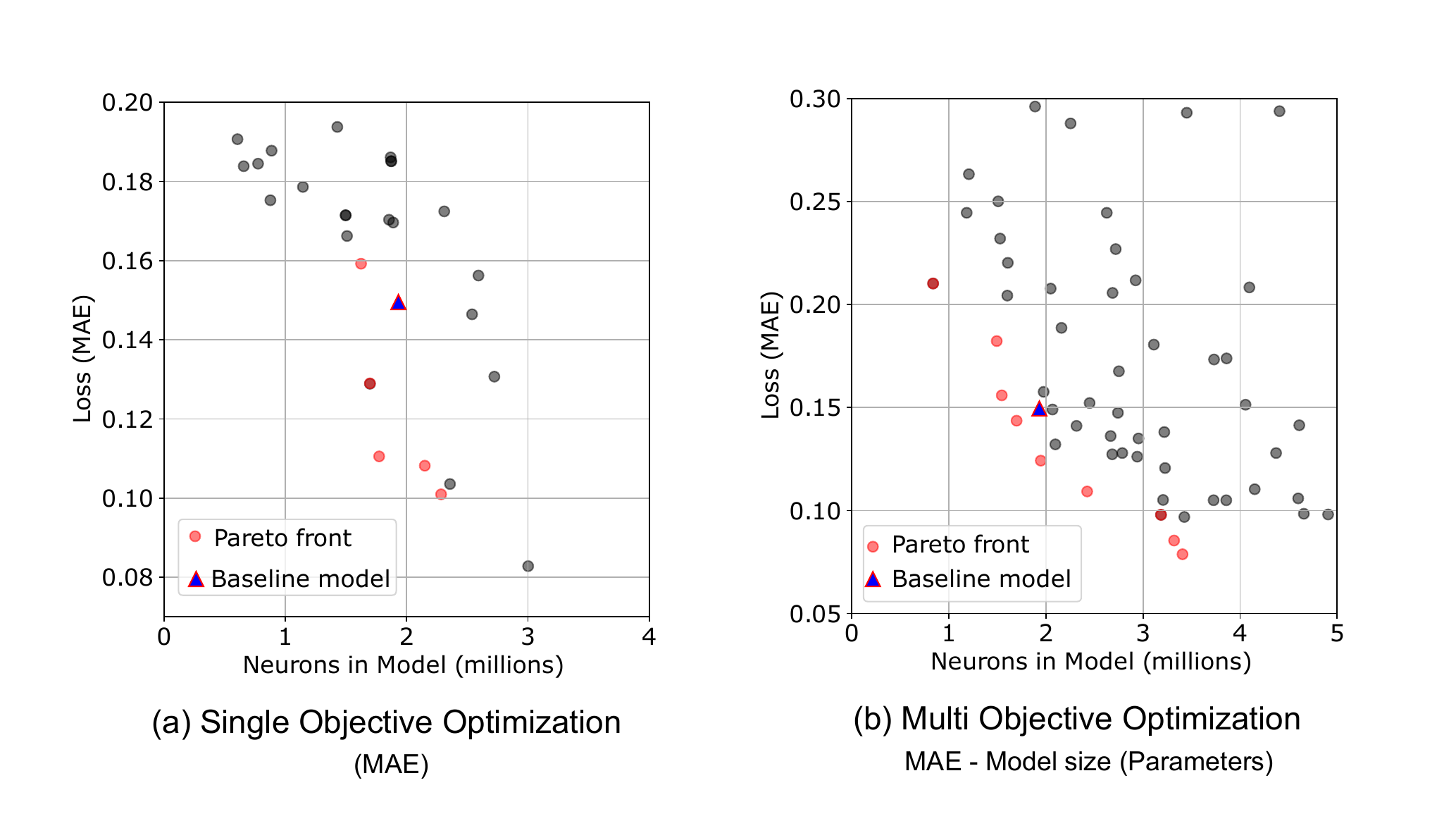}}
	\caption{PtychoNN Model: Pareto front of (a) single objective - model accuracy and (b) multi-objective optimization, respectively}
	\vspace{-15pt}
	\label{fig:ptychonnsw-pareto}
\end{figure*}

\subsection{Setup} \label{subsec:label}
We use the computing resources provided on Swing at the high-performance computing (HPC) cluster operated by the Laboratory Computing Resource Center (LCRC) at Argonne National Laboratory. It consists of 6 nodes, each operating 2$\times$ AMD EPYC 7742 64-Core Processors and 8$\times$ NVIDIA A100 GPUs with 320GB of GPU memory, 1TB of DDR4 memory and 14TB of local scratch. 

\subsection{Inference Platform} \label{subsec:platform}
Due to the nature of the applications, edge AI accelerators are highly suitable for providing direct and rapid feedback/control to sensors at a low cost. Therefore, we a popular edge computing platform: the Nvidia Jetson AGX Orin 32GB (Orin-AGX) module. The Orin-AGX features a 1792-core Ampere architecture GPU with 56 Tensor Cores, achieving 200 TOPS. As for the CPU host, Orin-AGX is supported by an 8-core Arm Cortex-A78AE. The models in the paper are entirely executed on the Orin-AGX module; thus, the CPU do not significantly contribute to the performance differences across models. Regarding power consumption, the Orin-AGX's consumption ranges from 15 to 40W. It is possible to limit power consumption with configuration, but we have not pursued this option.


We utilized the Nvidia's optimized inference procedure recommended by the official Nvidia tool, TensorRT. Initially, we created optimized inference engines for each batch size using INT8 quantization. Subsequently, these engines were employed to benchmark the latency of each model. Each model was executed 375 times, each with 10 iterations, and the average latency is reported. It's important to note that although Orin-AGX is equipped with Nvidia's custom DL accelerator, NVDLA, the DLA cores are not utilized by TensorRT. For power measurement, we used the Jetson power rail pins to report the total power/energy at the module level for each inference.




\section{Deephyper Results and Discussion}\label{sec:results}
In this section, we evaluate the optimized BraggNN and PtychoNN models generated by the automated NAS and HPS. We compare the generated models' \textit{accuracy}, \textit{size} (trainable parameters), and inference \textit{latency} and \textit{energy consumption}, with their respective baseline models.

\subsection{Single-Objective Optimization - Accuracy}
In this section, we investigate the baseline BraggNN and PtychoNN models to optimize a single-objective, application accuracy.

\begin{figure*}[h!]
	\centering
	\centerline{\includegraphics[height=0.9\columnwidth, width=0.95\textwidth]{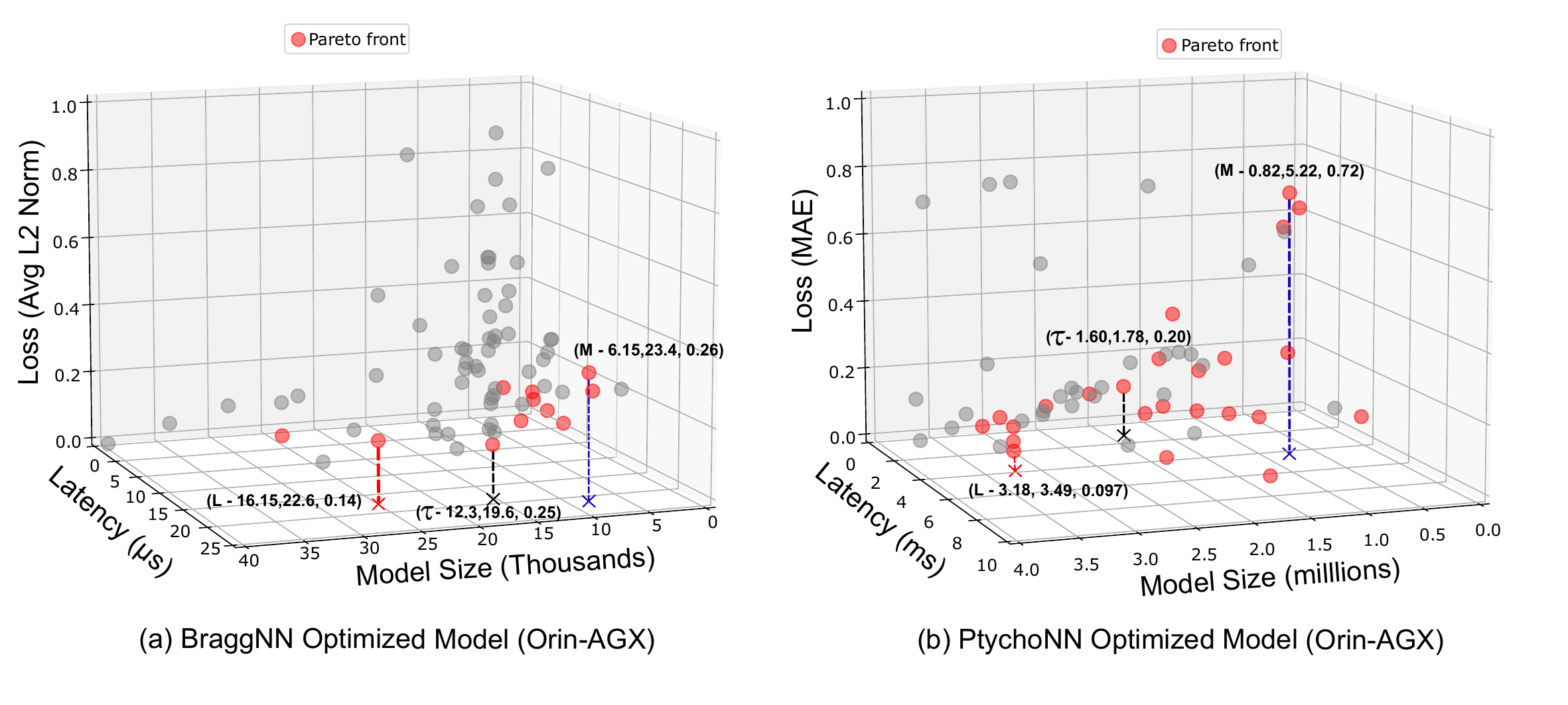}}
	\caption{HW Inference Performance: (a) BraggNN - Pareto front of optimized models and their inference latency (ms), and (b) PtychoNN: Pareto front of optimized models and their inference latency (ms). Results values: (Model Size(M), Latency($\tau$), Loss(L)) }
	\label{fig:hw-Pareto}
\end{figure*}

\subsubsection{BraggNN} 
We explore the network architecture and the hyperparameters, described in section \ref{sec:modelopt}, to minimize the mean square error (MSE) of the predicted bragg peak detection. Figure \ref{fig:braggsw-pareto} (a) illustrates the Pareto optimal models for bragg peak detection and compares its performance, in terms of its average MSE loss, with the size of the optimized model.

We observe that the \textit{mlpbragg} outperforms the baseline model in terms of accuracy. The mlpbragg model has an average MSE loss of 0.21, a 31.3\% improvement and a model size of 6,150 trainable parameters, an 87.57\% reduction in model size when compared to the baseline BraggNN model.

\subsubsection{PtychoNN} 
We explore the network architecture and the hyperparameters, described in section \ref{sec:modelopt}, to minimize the mean absolute error (MAE) of the predicted real-space amplitude and phase. Figure \ref{fig:ptychonnsw-pareto} (a) illustrates the Pareto optimal models for ptychographic reconstruction and compares its performance, in terms of its average MAE loss, with the size of the optimized model.

We observe that the \textit{cnnPtycho} outperforms the baseline model in terms of accuracy. The optimal cnnPtycho model has an average MAE loss of 0.12, a 16.77\% improvement and a model size of 1,756,800 trainable parameters, an 12.67\% reduction in model size when compared to the baseline PtychoNN model. 

\subsection{Multi-objective Optimization} \label{subsec:multi}
In this section, we investigate the multi-objective optimization of the BraggNN and PtychoNN model using Deephyper.

\subsubsection{BraggNN} 
We explore the network architecture and the hyperparameters, described in section \ref{sec:modelopt}, to minimize the mean square error (MSE) of the predicted bragg peak detection. Figure \ref{fig:braggsw-pareto} (b) illustrates the Pareto optimal models highlights the relation between the chosen objectives - model accuracy, quantile loss and model size.

We observe that the \textit{mlpbragg} outperforms the baseline model in terms of accuracy, quantile loss and model size. The mlpbragg model with optimal accuracy is highlighted (L) in figure \ref{fig:braggsw-pareto} (b). The mlpbragg has an average MSE loss of 0.14, a 51.72\% improvement and a model size of 16,150 trainable parameters, an 67.37\% reduction in model size when compared to the baseline BraggNN model.

\subsubsection{PtychoNN} 
We explore the network architecture and the hyperparameters, described in section \ref{sec:modelopt}, to minimize the mean absolute error (MAE) and model size of the baseline PtychoNN model. Figure \ref{fig:ptychonnsw-pareto} (b) illustrates the Pareto optimal models for ptychographic reconstruction and compares the model performance (average MAE loss) with the optimized model size.

We observe that the \textit{cnnPtycho} outperforms the baseline model in terms of accuracy and model size. The optimal cnnPtycho model has an average MAE loss of 0.12, a 16.77\% improvement and a model size of 1,708,150 trainable parameters, an 12.82\% reduction in model size when compared to the baseline PtychoNN model. 


\begin{table}[h!]
    \centering
	\renewcommand{\arraystretch}{1.2}
	\setlength{\tabcolsep}{2pt}
	\caption{Inference latency ($\tau$) and energy consumption of the baseline and optimized benchmarks with 32 batch size.}
	\label{tab:hw-metrics} \resizebox{\columnwidth}{!}
 {%
		\begin{tabular}{|l|c|c|c|}
		\hline
		\textbf{Application} & \textbf{Latency (ms)} & \textbf{Energy Consumption (mJ)} \\
        \hline
        Baseline BraggNN & 0.19 & 4.38 \\
         \hline 
        Optimized BraggNN & 0.17 & 1.94 \\
        \hline  
        Baseline PtychoNN & 1.69 & 52.22 \\
         \hline 
        Optimized PtychoNN& 1.53 & 44.21 \\
        \hline 
        \end{tabular}}
\end{table}

\subsection{Hardware Evaluation}
In this section, we compare the performance of the optimized models and their respective baselines in terms of latency and energy consumption when executed on the Nvidia Jetson AGX Orin platform. We highlight the optimal models at the edge of each metric - Loss(L), Latency ($\tau$) and Model size (M). Table \ref{tab:hw-metrics} highlight the latency and energy consumption to infer the baseline and optimized models.

Figure \ref{fig:hw-Pareto} (a) illustrates the Pareto front in the multi-objective optimization of the bragg models discussed in Section \ref{subsec:multi}. We observe that the mlpbragg model outperforms the baseline in terms of energy consumption. The optimized mlpBragg demonstrates a 10.51\% and 44.18\% reduction in latency and energy consumption when compared to the baseline, when inferred on the Orin-AGX platform. Figure \ref{fig:hw-Pareto} (b) illustrates the Pareto front in the multi-objective optimization of the ptychoNN models. Similarly, we observe that the cnnptycho model outperforms the baseline in terms of energy consumption. The optimized cnnptycho has an 9.47\% and 18.79\% reduction in latency and energy consumption when compared to the baseline, when inferred on the Orin-AGX platform.





\section{Conclusion}
In this work, we explore neural architecture search (NAS) and Hyper-parameter search techniques to optimize neural network models for X-ray and electron diffraction-based microscopy using bragg peak detection and ptychography to perform 3-D imaging at an atomic resolution. We automate the design and optimization of the ANN models for models size (trainable parameters), energy consumption and throughput when evaluated on the Orin AGX platform. Typically, these techniques are implemented using computationally complex task such as a Psuedo-Voigt function or solving a complex inverse problem. We demonstrate the improved performance of the auto-tuned models when compared to the manually tuned BraggNN and PtychoNN benchmarks. We study and demonstrate the importance of the exploring the search space of tunable hyperparameters in enhancing the performance of bragg peak detection and ptychographic reconstruction. Our NAS and HPS of achieves a 31.03\% improvement in bragg peak detection accuracy with a 87.57\% reduction in model size. We also achieve a 16.77\% improvement in predicting the amplitude and phase for the ptychographic reconstruction expriment with a 12.82\% reduction in model size. When inferred on the Orin-AGX platform, the optimized Braggnn and Ptychonn models demonstrate a 10.51\% and 9.47\% reduction in inference latency and a 44.18\% and 15.34\% reduction in energy consumption when compared to their respective baselines, when inferred in the Orin-AGX edge platform.

\bibliography{references, dhyper}

\end{document}